\documentclass{mic2015}

\usepackage[sort&compress,numbers]{natbib}
\usepackage{amsmath}
\usepackage{listings}
\usepackage{xspace}
\usepackage{color}
\usepackage[colorlinks=false,allbordercolors={1 1 1}]{hyperref}

\newcommand{\train}{\mathbf{X}^{(tr)}}
\newcommand{\test}{\mathbf{X}^{(te)}}
\newcommand{\obj}{\mathcal{F}}
\DeclareMathOperator*{\argmin}{arg\,min}

\definecolor{mygreen}{rgb}{0,0.6,0}
\definecolor{mygray}{rgb}{0.5,0.5,0.5}
\definecolor{bggray}{rgb}{0.95,0.95,0.95}
\definecolor{mymauve}{rgb}{0.58,0,0.82}
\lstdefinestyle{Py}{
    language={Python}, 
    moredelim=**[is][\slshape\bfseries]{`}{`},
    moredelim=**[is][\color{mygreen}]{!}{!},
}
\begin{document}

\title{Hyperparameter Search in Machine Learning}
%\author{Marc Claesen\inst{1} \and Bart De Moor\inst{1}}
\author{Marc Claesen \and Bart De Moor}
\institute{
STADIUS Center for Dynamical Systems, Signal Processing and Data Analytics \\ 
KU Leuven, Department of Electrical Engineering (ESAT) -- iMinds, Department of Medical IT \\
Kasteelpark Arenberg 10, box 2446, 3001 Leuven, Belgium \\
\email{marc.claesen@esat.kuleuven.be, bart.demoor@esat.kuleuven.be}
}
\id{14}
\maketitle

\begin{abstract}
We describe the hyperparameter search problem in the field of machine learning and discuss its main challenges from an optimization perspective. Machine learning methods attempt to build models that capture some element of interest based on given data. Most common learning algorithms feature a set of hyperparameters that must be determined before training commences. The choice of hyperparameters can significantly affect the resulting model's performance, but determining good values can be complex; hence a disciplined, theoretically sound search strategy is essential.
\end{abstract}

\section{Introduction}
Machine learning research focuses on the development of methods that are capable of capturing some element of interest from a given data set. Such elements include but are not limited to coherent structures within data (clustering) or the ability to predict certain target values based on given characteristics, which may be discrete (classification) or continuous (regression). %For simplicity, we will assume machine learning approaches generally follow two phases: (i) training a model based on a given data set and (ii) testing the model on unseen data; though this division does not always exist \citep{joachims2003transductive}.

A large variety of learning methods exist, ranging from biologically inspired neural networks \citep{bishop1995neural} over kernel methods \citep{scholkopf2002learning} to ensemble models \citep{breiman2001random, JMLR:v15:claesen14a}. A common trait in these methods is that they are parameterized by a set of hyperparameters $\lambda$, which must be set appropriately by the user to maximize the usefulness of the learning approach. Hyperparameters are used to configure various aspects of the learning algorithm and can have wildly varying effects on the resulting model and its performance. 

Hyperparameter search is commonly performed manually, via rules-of-thumb \citep{hsu2003practical,hinton2012practical} or by testing sets of hyperparameters on a predefined grid \citep{pedregosa2011scikit}. These approaches leave much to be desired in terms of reproducibility and are impractical when the number of hyperparameters is large \citep{DBLP:journals/corr/ClaesenSPMM14}. Due to these flaws, the idea of automating hyperparameter search is receiving increasing amounts of attention in  machine learning, for instance via benchmarking suites \citep{eggensperger2013towards} and various initiatives.\footnote{Such as \url{http://www.automl.org/} and \url{https://www.codalab.org/competitions/2321}.} %the ChaLearn AutoML challenge \citep{guyondesign}.
Automated approaches have already been shown to outperform manual search by experts on several problems \citep{ bergstra2011algorithms,bergstra2012random}. 

We briefly introduce some key challenges inherent to hyperparameter search in Section~\ref{challenges}. The combination of all these hurdles make hyperparameter search a formidable optimization task. In Section~\ref{state-of-the-art} we give a succinct overview of the current state-of-the-art in terms of algorithms and available software.

\subsection{Example: controlling model complexity}
A key balancing act in machine learning is choosing an appropriate level of model complexity: if the model is too complex, it will fit the data used to construct the model very well but generalize poorly to unseen data (overfitting); if the complexity is too low the model won't capture all the information in the data (underfitting). This is often referred to as the bias-variance trade-off \citep{geman1992neural, cucker2002best}, since a complex model exhibits large variance while an overly simple one is strongly biased. Most general-purpose methods feature hyperparameters to control this trade-off; for instance via regularization as in support vector machines and regularization networks \citep{evgeniou2000regularization,hastie2004entire}.

\subsection{Formalizing hyperparameter search}
The goal of many machine learning tasks can be summarized as training a model $\mathcal{M}$ which minimizes some predefined loss function $\mathcal{L}(\test;\ \mathcal{M})$ on given test data $\test$. Common loss functions include mean squared error and error rate. The model $\mathcal{M}$ is constructed by a learning algorithm $\mathcal{A}$ using a training set $\train$; typically involving solving some (convex) optimization problem. The learning algorithm $\mathcal{A}$ may itself be parameterized by a set of hyperparameters $\lambda$, e.g. $\mathcal{M} = \mathcal{A}(\train;\ \lambda)$. An example model $\mathcal{M}$ is a support vector machine classifier with Gaussian kernel \citep{scholkopf2002learning}, for which the training problem $\mathcal{A}$ is parameterized by the regularization constant $C$ and kernel bandwidth $\sigma$, i.e. $\lambda = [C,\sigma]$.

The goal of hyperparameter search is to find a set of hyperparameters $\lambda^\star$ that yield an optimal model $\mathcal{M}^\star$ which minimizes $\mathcal{L}(\test;\ \mathcal{M})$. This can be formalized as follows \citep{DBLP:journals/corr/ClaesenSPMM14}:
\begin{equation}
\lambda^\star = \argmin_{\lambda} \mathcal{L}\big(\test;\ \mathcal{A}(\train;\ \lambda)\big) = \argmin_{\lambda} \obj(\lambda;\ \mathcal{A},\ \train, \test,\ \mathcal{L}). \label{equation}
\end{equation}
The objective function $\obj$ takes a tuple of hyperparameters $\lambda$ and returns the associated loss. The data sets $\train$ and $\test$ are given and the learning algorithm $\mathcal{A}$ and loss function $\mathcal{L}$ are chosen . Depending on the learning task, $\train$ and $\test$ may be labeled and/or equal to each other. In supervised learning, a data set is often split into $\train$ and $\test$ using hold-out or cross-validation methods \citep{efron1983leisurely,kohavi1995study}.

\section{Challenges in hyperparameter search} \label{challenges}
The characteristics of the search problem depend on the learning algorithm $\mathcal{A}$, the chosen loss function $\mathcal{L}$ and the data set $\train$, $\test$, as shown in Equation~\eqref{equation}. Hyperparameter search is typically approached as a non-differentiable, single-objective optimization problem over a mixed-type, constrained domain. In this section we will discuss the origins and consequences of challenges in hyperparameter search.

\subsection{Costly objective function evaluations} \label{time}
Each objective function evaluation requires evaluating the performance of a model trained with hyperparameters $\lambda$. Depending on the available computational resources, the nature of the learning algortihm $\mathcal{A}$ and size of the problem ($\train$, $\test$) each evaluation may take considerable time. Training times in the order of minutes are considered fast, since days and even weeks are not unheard of \citep{krizhevsky2012imagenet, dean2012large, sutskever2014sequence}. Evaluation time is exacerbated when procedures that train multiple models are employed; for instance to reliably estimate generalization performance \citep{efron1983leisurely,kohavi1995study}. This leads to an increasing need for efficient methods to optimize hyperparameters that require a minimal amount of objective function evaluations.

Additionally, the time required to train and test models can be contingent upon the choice of hyperparameters. Some hyperparameters have an obvious influence on train and/or test time, e.g. the architecture of neural networks \citep{bishop1995neural} and size of ensembles \citep{breiman2001random, JMLR:v15:claesen14a}. The influence of hyperparameters can also be subtle, for instance regularization and kernel complexity can significantly affect training time for support vector machines \citep{bottou2007support}.

\subsection{Randomness} \label{randomness}
The objective function often exhibits a stochastic component, which can be induced by various components of the machine learning pipeline, for example due to inherent randomness of the learning algorithm (initialization of a neural network, resampling in ensemble approaches, \ldots) or due to finite sample effects in estimating generalization performance. This stochasticity can sometimes be addressed via machine learning techniques; but unfortunately such solutions typically dramatically increase the time required per objective function evaluation, limiting their usefulness in some settings.

This inherent stochasticity directly implies that the empirical best hyperparameter tuple, obtained after a given set of evaluations, is not necessarily the true optimum of interest $\lambda^\star$. Fortunately, many search methods are designed to probe many tuples close to the empirical best. If the search region surrounding the empirical optimum is densely sampled, we can determine whether the empirical best was an outlier or not in a post-processing phase, for instance by assuming Lipschitz continuity or smoothness.

\subsection{Complex search spaces}
The number of hyperparameters is usually small ($\leq5$), but it can range up to hundreds for complex learning algorithms \citep{bergstra2013making} or when preprocessing steps are also subjected to optimization \citep{hutter2009paramils}. It has been demonstrated empirically that in many cases only a handful of hyperparameters significantly impact performance, though identifying the relevant ones in advance is difficult \citep{bergstra2012random}.

Hyperparameters are usually of continuous or integer type, leading to mixed-type optimization problems. Continuous hyperparameters are commonly related to regularization.  Common integer hyperparameters are related to network architecture for neural networks \citep{bishop1995neural}, size of ensembles \citep{breiman2001random, JMLR:v15:claesen14a} or the parameterization of kernels in kernel methods \citep{scholkopf2002learning}.

Some tasks feature highly complex search spaces, in which the very existence of certain hyperparameters are conditional upon the value of others \citep{hutter2009paramils,bergstra2011algorithms,bergstra2013hyperopt}. A simple example is optimizing the architecture of neural networks \citep{bishop1995neural}, where the number of hidden layers is one hyperparameter and the size of each layer induces a set of additional hyperparameters, conditional upon the number of layers. 

\section{Current approaches} \label{state-of-the-art}
A wide variety of optimization methods have been used for hyperparameter search, including particle swarm optimization \citep{meissner2006optimized, lin2008particle}, genetic algorithms \citep{tsai2006tuning}, coupled simulated annealing \citep{xavier2010coupled} and racing algorithms \citep{birattari2010f}. Surprisingly, randomly sampling the search space was only established recently as a baseline for comparison of optimization methods \citep{bergstra2012random}. Bayesian and related sequential model based optimization techniques using variants of the expected improvement criterion \citep{jones1998efficient} are receiving a lot of attention currently \citep{bergstra2011algorithms, hutter2011sequential, snoek2012practical, bachoc2013cross, eggensperger2013towards}, owing to their efficiency in terms of objective function evaluations.

Software packages are being released which implement various dedicated optimization methods for hyperparameter search. Such packages are usually intended to be used in synergy with machine learning libraries that provide learning algorithms \citep{pedregosa2011scikit}. Most of these packages focus on Bayesian methods \citep{hutter2009paramils, snoek2012practical, bergstra2013hyperopt}, though metaheuristic optimization approaches are also offered \citep{DBLP:journals/corr/ClaesenSPMM14}. The increased development of such packages testifies towards the growing interest in automated hyperparameter search.

\section{Conclusion}
A fully automated, self-configuring learning strategy can be considered the holy grail of machine learning. Though the current state-of-the-art still has a long way to go before this goal can be reached, it is evident that hyperparameter search is a crucial element in its pursuit. Automated hyperparameter search is a hot topic within the machine learning community which we believe can benefit greatly from the techniques and lessons learnt in metaheuristic optimization.

\section*{Acknowledgments}
This research was funded by the Flemisch Government via the following funding channels: FWO: projects:  G.0871.12N (Neural circuits); IWT: TBM-Logic Insulin(100793), TBM Rectal Cancer(100783), TBM IETA(130256), PhD grants (111065); Industrial Research fund (IOF): IOF Fellowship 13-0260; iMinds Medical Information Technologies SBO 2015, ICON projects (MSIpad, MyHealthData); VLK Stichting E. van der Schueren: rectal cancer and the Belgian Federal Government via FOD: Cancer Plan 2012-2015 KPC-29-023 (prostate) and COST: Action: BM1104: Mass Spectrometry Imaging.

%\begin{itemize}
%\setlength\itemsep{0.2em}
%\item Research Council KU Leuven: GOA/10/09 MaNet, CoE PFV/10/016 SymBioSys; 
%\item Flemish Government: FWO: projects:  G.0871.12N (Neural circuits); IWT: TBM-Logic Insulin(100793), TBM Rectal Cancer(100783), TBM IETA(130256), O\&O ExaScience Life Pharma, ChemBioBridge, PhD grants (specifically 111065); Industrial Research fund (IOF): IOF/HB/13/027 Logic Insulin; iMinds Medical Information Technologies SBO 2014%; VLK Stichting E. van der Schueren: rectal cancer
%\item Federal Government: FOD: Cancer Plan 2012-2015 KPC-29-023 (prostate)
%\item COST: Action: BM1104: Mass Spectrometry Imaging
%\end{itemize}

\newpage
\bibliography{paper}
\bibliographystyle{plain}

\end{document}